\documentclass{article}
\usepackage{ijcai25}

\usepackage{times}
\usepackage{soul}
\usepackage{url}
\usepackage[hidelinks]{hyperref}
\usepackage[utf8]{inputenc}
\usepackage[small]{caption}
\usepackage{graphicx}
\usepackage{amsmath}
\usepackage{amsthm}
\usepackage{booktabs}
\usepackage{algorithm}
\usepackage{algorithmic}
\usepackage[switch]{lineno}
\usepackage{multirow}
\usepackage{xcolor}

\title{ChronoFact: Timeline-based Temporal Fact Verification}

\author{
Anab Maulana Barik$^1$ \and
Wynne Hsu$^{1,2}$\And
Mong Li Lee$^{1,3}$\\
\affiliations
$^{1}$School of Computing,
$^{2}$Institute of Data Science,
$^{3}$Centre for Trusted Internet and Community \\
National University of Singapore, Singapore \\
\emails
anabmaulana@u.nus.edu,
\{whsu,leeml\}@comp.nus.edu.sg
}

\begin{document}

\maketitle

\begin{abstract}
Temporal claims, often riddled with inaccuracies, are a significant challenge in the digital misinformation landscape. Fact-checking systems that can accurately verify such claims are crucial for combating misinformation. Current systems  struggle with the complexities of evaluating the accuracy of these claims, especially when they include multiple, overlapping, or recurring events.
We introduce a novel timeline-based fact verification framework that identify events from both 
 claim and evidence and organize them into their respective chronological timelines. 
 The framework systematically examines the relationships between the events in both claim and evidence  to predict the veracity of each claim event and their chronological accuracy. This  allows us to accurately determine the overall veracity of the claim. 
 We also introduce a new dataset  of complex temporal claims involving  timeline-based reasoning for  the training and evaluation of our proposed framework. Experimental results demonstrate the effectiveness of our approach in handling the intricacies of temporal claim verification.
\end{abstract}

\section{Introduction}
The spread of false information has reached alarming levels, undermining social trust  significantly.  One prominent category of false information involves temporal claims, which are statements that include 
time-specific elements, either explicitly (e.g., "in 1953") or implicitly (e.g., "before [another event]"). 
The complexity of verifying these claims escalates with the number of events mentioned. Effective verification must assess not only the veracity of each event within its temporal context but also understand the relationships between these events, especially their chronological order.

Existing research has largely focused on verifying the veracity of individual events within claims, and overlook the chronological order of these events \cite{tacv,temporalfc}. 
This oversight can undermine the effectiveness of fact-checking systems, particularly when dealing with complex narratives where the sequence of events is crucial for determining the truth.
Understanding the chronological order of events is hindered by the absence of the explicit temporal cues. The complexity increases when events overlap or recur, complicating the task of establishing a clear and accurate timeline for the verification process.

\begin{figure}[t!] 
\centering
\includegraphics[width=0.48\textwidth]{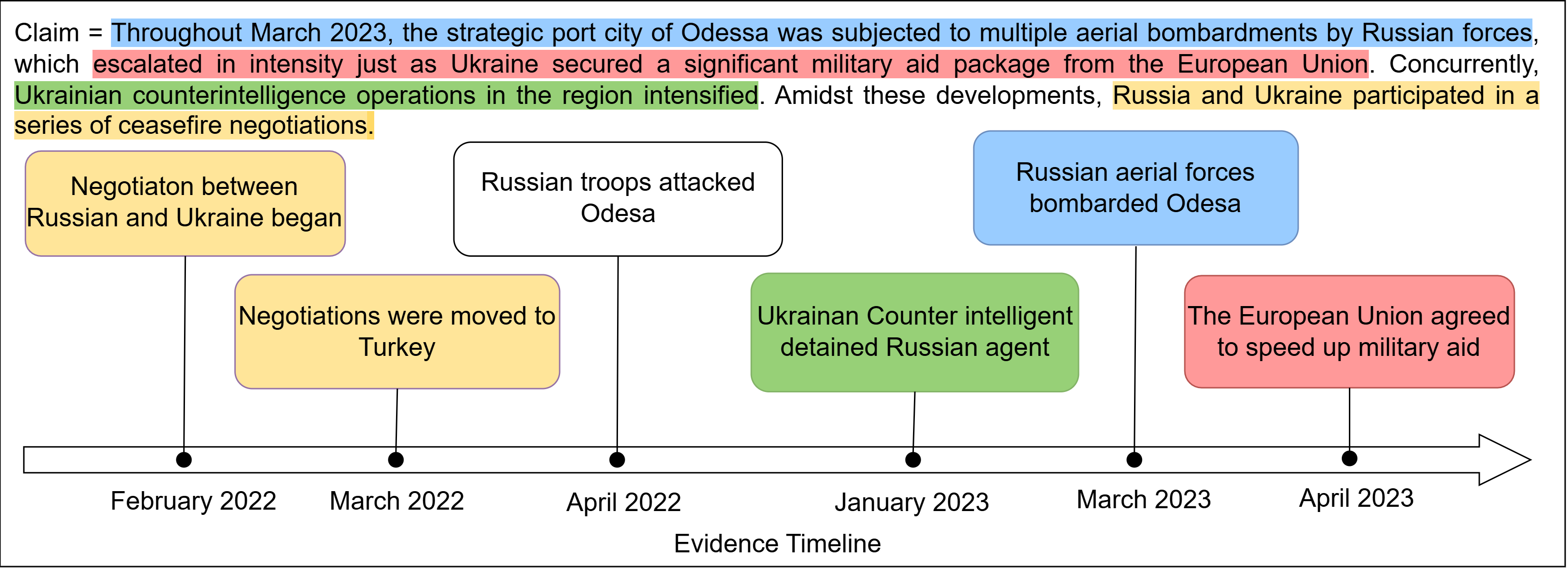} 

(a) Claim with multiple events\\ \smallskip

\includegraphics[width=0.48\textwidth]{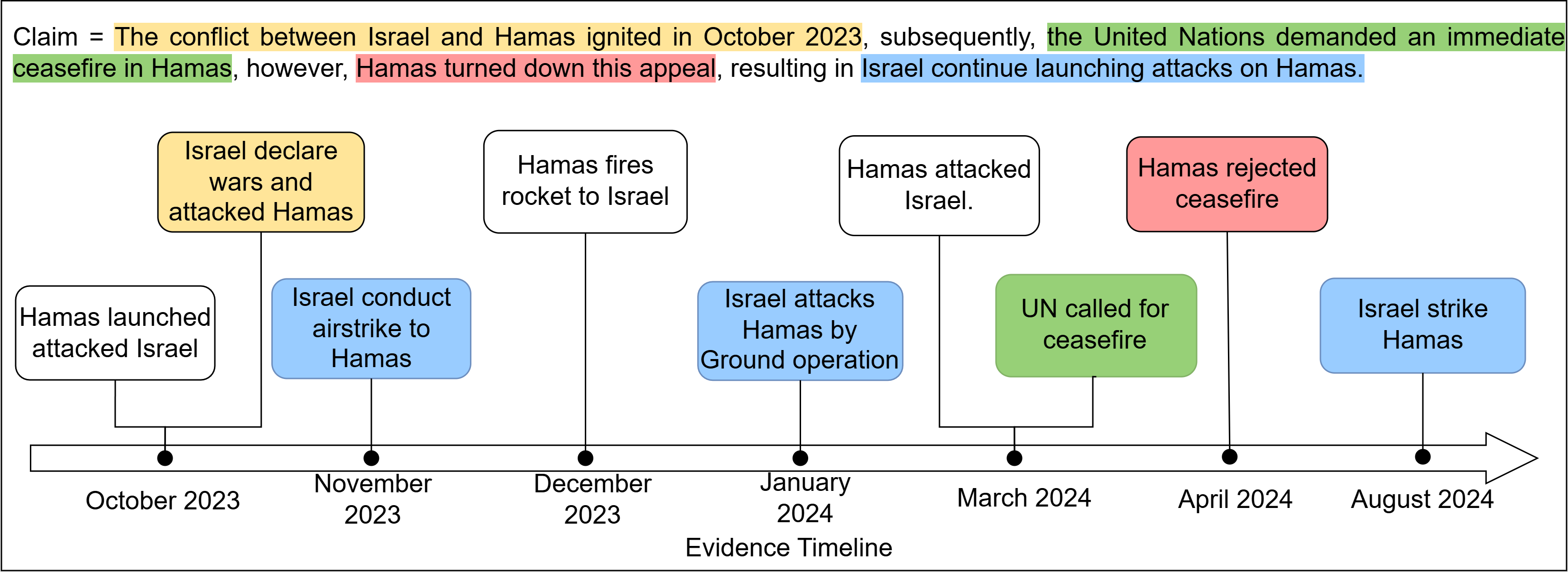}

(b) Claim with recurring event
\vspace*{-0.1in}
\caption{Illustration of complex temporal claim verification.}
\label{fig:example_4}
\vspace*{-0.17in}
\end{figure}

\smallskip
\noindent \textbf{Example 1.} 
 Figure~\ref{fig:example_4}(a) shows a claim involving four events, with
 evidence relevant to each event  indicated by matching color boxes. While each event appears to be supported by some evidence when analyzed independently, closer scrutiny of the timeline reveals discrepancies. 
For instance, evidence indicates that negotiations between Ukraine and Russia took place in February 2022, before the bombardments in March 2023. Therefore, the claim event \textit{"Russia and Ukraine participated in a series of ceasefire negotiations"}, suggested by the temporal cues to have occurred in March 2023, is not actually supported by the evidence.
Existing works that analyze individual events without considering the timeline  would incorrectly conclude that  all the events in the claim are supported, and deem the claim true.
This shows the importance of  \textbf{chronological order of events} in claim verification.

\begin{figure*}[t!] 
\centering
\includegraphics[width=.8\textwidth]{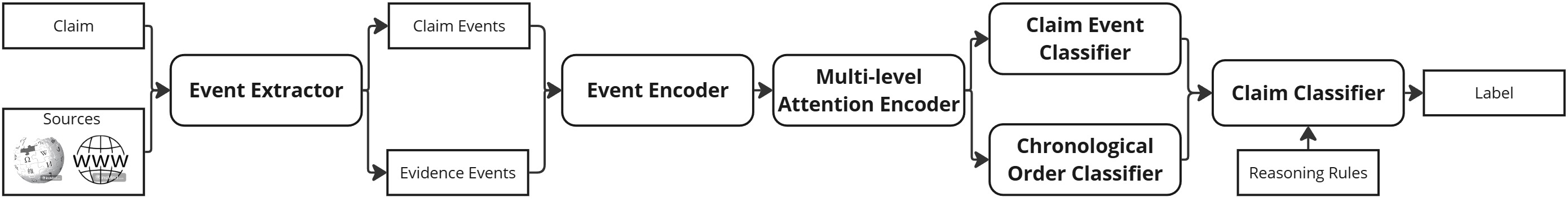}
\caption{Overview of \texttt{ChronoFact} Framework.}
\label{fig:framework}
\vspace*{-0.2in}
\end{figure*}

\smallskip
\noindent \textbf{Example 2.} 
 Figure~\ref{fig:example_4}(b) shows a
 claim event \textit{''Israel attacks Hamas''}, with evidence indicating  it occurs at three different times:  November 2023, January 2024, and August 2024.
 Analysis of the timeline shows that the occurrence in August 2024  aligns with the order of events in the claim, thereby concluding that the claim is supported. 
  Previous studies that have overlooked the timeline of events might match the claim event to earlier occurrence in Nov 2023, or Jan 2024, leading to  incorrect conclusion that this claim is refuted.
 This highlights the need to \textbf{align claim and evidence events with their  chronological order} for accurate claim verification.
 
 \smallskip
To overcome the limitations in existing temporal fact verification methods, we introduce \texttt{ChronoFact}, a framework that  systematically identify events from both claim and evidence to construct a coherent timeline for temporal fact verification.
\texttt{ChronoFact} examines the relationships between  claim and evidence events  at three levels: event-level, token-level, and time-level and learns
 a model to predict the veracity of  each claim event  and their chronological accuracy. 
 
We develop a new dataset called \texttt{ChronoClaims} for timeline-based fact verification. This  dataset 
encompasses complex claims involving multiple, related events that unfold over time with both 
implicit and explicit  temporal expressions.
 These features are lacking in  
current datasets like T-FEVER, which mainly consist of single-event claims, and T-FEVEROUS, which, despite its more complex claims, often has events that are not chronologically related. 

\section{Related Work}
There has been a stream of research on evidence-based claim verification to assess
whether the evidence sentences support or refute the claim \cite{domlin,bert-fact-checking}.  Techniques like  GEAR \cite{gear}, KGAT \cite{KGAT}, and DREAM \cite{DREAM}  model  claim and evidence into a graph using Graph Attention Network to facilitate information propagation of information between them.
CGAT \cite{cgat} enhances this by incorporating commonsense knowledge from ConceptNet to enrich the contextual representation. These works do not consider temporal information.

Only a few studies have incorporated temporal information in the claim verification process. NMT \cite{neural-mt-temporal-claims}  
verifies economic claims against time series data in tabular format by
 translating the claims into Datalog rules to query the tabular evidence.
TemporalFC \cite{temporalfc} focuses on verifying claims represented as  tuples  against a knowledge graph. 
It utilizes temporal graph embeddings to determine the validity timing of underlying triples. 
However, these works are limited to structured data, hindering its applicability to natural language claims and evidence.
 
\cite{time-aware-reranking} capture the temporal relevance of evidence through a re-ranking process based on the proximity of publication dates between the evidence and the claim.
ITR \cite{implicit-temporal-reasoning} extends this by assigning publication dates of claims and evidence to fixed-size time buckets for temporal reasoning. 
However, it ignores implicit temporal cues or chronological order in the claim and evidence.

TACV \cite{tacv} decomposes claim and evidence sentences into events and employ temporal-aware representation encoder to retrieve evidence that are both semantically and temporally related to the claim.
It utilizes GPT  for temporal reasoning to verify individual claim events.
Unlike TACV, our approach also evaluates the chronological timeline accuracy between the claim and evidence, providing a more thorough verification of complex claims.

A related area of research, question answering (QA), has also begun to incorporate temporal information.
FAITH \cite{faith} focuses on implicit temporal reasoning for the temporal question answering task by generating intermediate questions with explicit temporal information.
This method uses heterogeneous sources to improve the completeness of evidence retrieval and employs a general-purpose QA system to respond to the question.
While it is common to utilize QA in the claim verification process \cite{qacheck}, they do not consider the chronological order of events.
In contrast,  our work focuses on verifying complex claims that require timeline-based reasoning.

\section{Proposed Framework}

\texttt{ChronoFact} follows the typical automated claim verification which involves  collecting relevant evidence from credible sources and assessing the claim's veracity based on the evidence.
 \autoref{fig:framework} shows the key modules in  the framework.

\smallskip
\noindent\textbf{Event Extractor.} Given a claim, we first employ the GENRE sequence-to-sequence entity linking model \cite{genre} to retrieve relevant documents from Wikipedia and extract all evidence sentences.
Then we utilize Semantic Role Labelling from AllenNLP \cite{allennlp} to extract events from both the claim and evidence. Each event has its core information and temporal argument. 
Finally, we  score the evidence with the events extracted from the claim using an event representation encoder similar to \cite{tacv}.

\smallskip
\noindent\textbf{Event Encoder.} We tokenize each event and pass each token $i$ to the flan-T5 model to obtain the corresponding token representations $H_i$. 
 For tokens that represent date, we apply mean pooling, followed by positional encoding \cite{transformer} where the position corresponds to the  distance between the event date and the earliest date found in the claim and evidence events. 
The final representation of the event is given by $<$$H_{CLS}$, $H_1 \cdots H_d$$>$  where $d$ is the number of tokens in the event, $H_{CLS}$ is the event-level representation obtained by the average pooling of  $H_j, 1 \leq j \leq d$, and $H_1 ... H_d$ are the token-level representations.

\smallskip
\noindent\textbf{Multi-level Attention Encoder.}
We process  each pair of claim event representation $c_i$ and evidence event representation $e_j$ 
 through a multi-level attention module to 
 determine the relevance of evidence events for each claim event. This involves calculating attention scores at three  levels:

$\bullet$ \textbf{token-level attention score $\alpha_{ij}$}: is the average of  cosine similarities between all pairs of tokens in $c_i$ and $e_j$. 

$\bullet$ \textbf{event-level attention score  $\beta_{ij}$}: is derived from the   cosine similarity between the $H_{CLS}$ representations of claim $c_i$ and evidence $e_j$.
  
$\bullet$ \textbf{time-level attention score $\gamma_{ij}$}: is computed from  the cosine similarity between the mean pooled date representations in $c_i$ and $e_j$.

The  final multi-level attention score between $c_i$ and $e_j$,
denoted as $\omega_{ij}$, 
is the average of the event-level, token-level, and time-level attention scores.
 These computed  attention scores are then employed to predict the label of each event in the claim, assess the  accuracy of the claim’s chronological order based on the evidence timeline, and evaluate the claim's overall veracity.
Note that  we considered dynamically learn the weights of each attention level.
However, our experiment indicated  that this approach did not improve performance, leading us to adopt the average weighting method instead.

\smallskip
\noindent\textbf{Claim Event Classifier.}
This module predicts the label of claim event $c_i$ using the top-$k$ evidence events $E' \subset E$ with the highest final attention scores. 
Let $H^{c_i}_{CLS}$ and $H^{e_j}_{CLS}$ be the representation of the $CLS$ token for $c_i$ and evidence event $e_j \in E'$ respectively.
We concatenate the representations weighted by the attention scores to obtain $u^{c_i}$
as follows:
$$u^{c_i} = H^{c_i}_{CLS} \oplus \omega_{i1} H^{e_1}_{CLS} \oplus \cdots \oplus \omega_{ik} H^{e_k}_{CLS}$$
This is then fed to
two fully connected layers followed by  softmax  to obtain the probability distribution $z^{c_i}$: 
$$z^{c_i}=\mathrm{softmax}(FC_2(\mathrm{ReLU}(FC_1(u^{c_i}))))$$
where
$z^{c_i}[0]$, $z^{c_i}[1]$, and $z^{c_i}[2]$ are the probabilities of the labels "SUP", "REF", and "NEI" 
respectively. 
The label $y^{c_i}$ for the claim event $c_i$ is assigned based on the highest probability among these.

\smallskip
\noindent\textbf{Chronological Order Classifier.}
This module  predicts the accuracy of the  chronological order of a claim $C$ using the  timeline of evidence events.
Given $n$ events in  $C$,
the relevance of each evidence event to $C$ is given by:
\begin{equation}
r^{e_j} = \tanh(\sum_{i=1}^n \omega_{i,j})
\label{eq:reweight}
\end{equation}

We use  $\tanh$ for its bounded output range of $[-1, 1]$, which aligns with our interpretation of relevance as a continuous spectrum ranging from negative to positive.
We sort the evidence events based on their relevance  to the claim and obtain the top-$k$ events with the highest score $r^{e_j}$. 
These top-$k$ evidence events, along with the claim events, are then input into GPT to be reordered according to their chronological sequence. Given the limited number $k$ of events, this reordering process requires minimal time and resources, making it practical for real-world applications.

 The reordered sequence of claim events, denoted as $seqC = H^{c_1}_{CLS} \oplus \cdots \oplus H^{c_n}_{CLS}$ is passed to a Bi-LSTM 
to capture and embed the chronological order into the model. 
Similarly, the reordered sequence of evidence events, weighted by their relevance scores is given by $$seqE = r^{e_1 }H^{e_1}_{CLS} \oplus \cdots \oplus r^{e_m}H^{e_m}_{CLS}$$ 
This sequence is passed to a second Bi-LSTM.
The outputs from the two Bi-LSTM, 
denoted as $o^{C}$ for claim events and $o^{E}$ for evidence events, 
are then fed into two fully connected layers followed by  softmax 
to obtain the distribution $z^o$ 
$$z^o=\mathrm{softmax}(FC_4(\mathrm{ReLU}(FC_3([o^C \oplus o^E]))))$$
where $z^o[0]$ is the probability that the chronological order of the claim events is supported by that of the evidence events, and $z^o[1]$ is the probability that the chronological order  refutes the claim's timeline. The output of the chronological order classifier  $y^{o}$ is the label with the highest probability.

\smallskip
\noindent\textbf{Claim  Classifier.} To predict the overall veracity of the claim, we concatenate $seqC$ and $seqE$, along with the distributions
$z^{c_1} \cdots z^{c_n}, z^{o}$,  and pass this vector to two fully connected layers followed by a softmax function to obtain the probability distribution $z$ that the claim is SUP, REF or NEI. The label with the highest probability is depicted as $y$.

\begin{table*}[t!]
\vspace*{-0.2in}
\begin{center}
\scriptsize
\begin{tabular}{|c|c|c|c|c|c|c|c|c|}
\hline
\multirow{6}{*}{Temporal Expression}  & \multicolumn{2}{|c|}{}  & \multicolumn{2}{c|}{Train Set } & \multicolumn{2}{c|}{Validation Set } & \multicolumn{2}{c|}{Test Set } \\\cline{4-9}
& \multicolumn{2}{|c|}{}  & Support & Refute & Support & Refute & Support & Refute  \\\hline

& \multirow{3}{*}{Explicit} & 3 events  & 3,561 & 2,954 & 355 & 318 & 397 & 315 \\ \cline{3-9}

&  & 4 events  & 3,574 & 3,376 & 354 & 282 & 359 & 262 \\\cline{3-9}

& & 5 events  & 3,597 & 3,084 & 266 & 198 & 323 & 218 \\\cline{2-9}

& \multirow{3}{*}{Implicit} & 3 events & 3,804 & 3,028 & 355 & 318 & 390 & 313 \\ \cline{3-9}

& & 4 events  & 3,453 & 3,306 & 355 & 278 & 361 & 263 \\\cline{3-9}

& & 5 events & 3,473 & 3,039 & 266 & 199 & 321 & 213 \\\hline
\hline

 \multirow{2}{*}{Temporal Category}&  \multicolumn{2}{|c|}{Overlapping events} &10,360  & 9,297 & 918 & 807 & 985 & 714 \\ \cline{2-9}

& \multicolumn{2}{|c|}{Recurring events} & 8,914 & 6,507 & 729 & 376 & 865 & 361 \\ \hline
\end{tabular}
\vspace*{-0.1in}
\caption{Characteristics of ChronoClaims dataset.}
\label{table:tlver-statistics}
\end{center}
\vspace*{-0.225in}
\end{table*}

\subsection{Model Training}
We train the model using two losses
$\mathcal{L}_{cross}$ and $\mathcal{L}_{soft}$. The first loss $\mathcal{L}_{cross}$ is defined as follows:
\begin{equation}
\mathcal{L}_{cross} = \sum_{i=1}^{n} F(g^{c_i}, z^{c_i}) + F(g^{o}, z^{o}) + F(g, z)
\label{eq:loss-1}
\end{equation}
where $F(.)$ is a cross-entropy function, $g^{c_i}$, $g^{o}$, and $g$ are the ground-truth labels for claim event $c_i$, chronological order, and overall claim label respectively.
 
The second loss ensures the consistency between the overall claim label, claim event labels, and chronological accuracy. In particular, 
 we apply a set of logic rules based on the outcomes of the individual claim events and their chronological alignment with the evidence. Specifically, a claim is deemed supported if all its associated claim events are supported and their chronological sequence matches that of the evidence events. This is expressed using first-order logic:
$$y^{c_1} \land \cdots \land y^{c_n} \land y^o \implies y$$
This logical expression states that the overall veracity $y$ of a claim is SUP if and only if each claim event $y^{c_i}$
  is supported and the chronological order $y^o$  is consistent with the evidence. On the other hand, if any one of the claim event is refuted, or the chronological order does not align, then $y$ is REF. Otherwise, $y$ is NEI.
We leverage G\"{o}del t-norm to soften the hard reasoning rules in the claim classifier, and obtain the differentiable distribution $z_{soft}$:
\begin{equation}
\begin{aligned}
& z_{soft}[0] = min(z^{c_1}[0], \cdots, z^{c_n}[0], z^{o}[0]) \\
& z_{soft}[1] = max(z^{c_1}[1], \cdots, z^{c_n}[1], z^{o}[1]) \\
& z_{soft}[2] = 1 - z_{soft}[0] - z_{soft}[1] 
\\
\end{aligned}
\label{eq:logical-constraint}
\end{equation}
With this, we define $\mathcal{L}_{soft}$ which ensures  the consistency of the overall claim label  with the  claim events and their chronological order as follows:
\begin{equation}
\mathcal{L}_{soft} = D_{KL}(z || z_{soft})
\label{eq:loss-2}
\end{equation}
where $D_{KL}$ is Kullback-Leibler divergence, which measures the difference between the
predicted overall claim label distribution and the distribution derived from the soft logic.

The final loss function combines the soft logic loss $\mathcal{L}_{soft}$ with standard cross-entropy loss $\mathcal{L}_{cross}$, enabling the model to be trained both with supervision and structured constraints:
 \begin{equation}
\mathcal{L} = (1 - \mu) \mathcal{L}_{cross} + \mu \mathcal{L}_{soft}
\label{eq:loss_final}
\end{equation}
where $\mu$ is the hyperparameter.
 
\section{ChronoClaims Dataset}

We introduce a new benchmark dataset called \texttt{ChronoClaims} that is  designed for enhancing the accuracy and complexity of timeline-based fact verification. Utilizing the November 2022 Wikidata snapshot \cite{wikidata}, we preprocess and extract facts in the format \textless{}subject, relation, object, time\_start, time\_end\textgreater{}.
To construct an evidence timeline, we organize all the facts that have the same subject in a chronological order.
Then we randomly  select $N$ facts from this timeline and use GPT to transform these facts into coherent  sentences. 

For generating sentences with implicit temporal information, we omit the time\_start and time\_end from the fact, and prompt GPT to craft sentences that subtly embed the temporal context.
Then we use a cloze-style  template\footnote{\textless{}sentence$_1$\textgreater{} and then \textless{}sentence$_2$\textgreater{} and then \textless{}sentence$_3$\textgreater{}} to synthesize the sentences into a claim that preserves the chronological order.
To generate claims that contradict the evidence timeline, we rearrange the order of these sentences.
Finally, we use GPT to refine and rephrase the 
synthetic claim to make it more natural and fluent. Each  claim is labeled as either SUP or REF, depending on whether it aligns or conflicts  with the timeline.

We evaluate the quality of the \texttt{ChronoClaims} dataset by sampling 500 claims, ensuring equal distribution across different temporal expressions and event complexities (16\% each type), and temporal categories (50\% each category) to achieve representative coverage. 
Two human annotators are tasked with determining the labels of the generated claims based on the ground truth evidence timeline. The agreement rates are high, with 96.8\% and 97\% of the labels assigned by the annotators matching the labels of the generated claims.

We analyze the claims where the generated labels are different from the  annotators, and discover that most of the discrepancies are due to  errors in rephrasing
For example, a claim "Nasrallah Peter Sfeir worked as a Catholic priest, and then Nasrallah Peter Sfeir was educated at Saint Joseph University," was  rephrased to "Nasrallah Peter Sfeir pursued his education at Saint Joseph University before becoming a Catholic priest," which altered the original chronological sequence.
 To rectify this, we perform a second  verification step using GPT to ensure that the semantic meanings of
  the original and rephrased claims remain consistent. Any claims where the meaning has been altered are discarded. 

 In total, we generated 40,249, 3,544 and 3,735 claims for the training, validation and test sets. 
  \autoref{table:tlver-statistics} shows the detailed  statistics of the \texttt{ChronoClaims} dataset.

\section{Performance Study}

\noindent\textbf{Datasets.} Besides the \texttt{ChronoClaims} dataset, we also use  the  T-FEVER and T-FEVEROUS datasets  \cite{tacv}. These datasets 
are derived from the benchmark fact verification datasets FEVER \cite{fever}, FEVER2.0 \cite{fever2} and FEVEROUS \cite{feverous} respectively such that the claims in T-FEVER and T-FEVEROUS contain temporal expressions. Each claim involves  a  maximum of 3 events and is labeled as SUP, REF and NEI.
We also evaluate our method on the T-QuanTemp, a subset of the QuanTemp \cite{numtemp} dataset focusing on real-world claims with temporal aspects.
T-QuanTemp comprises of temporal claims containing temporal expressions or tagged as as temporal aspects by QuanTemp dataset.
Table~\ref{table:statistics2} shows the characteristics of  T-FEVER, T-FEVEROUS and T-QuanTemp datasets.
We use 80\% of the data for training and 20\% for testing.
While T-QuanTemp comes with its own set of evidence, we rely on different knowledge sources for the others: Wikipedia for T-FEVER and T-FEVEROUS, and Wikidata for ChronoClaims.

\begin{table}[h!]
\begin{center}
\scriptsize
\vspace*{-0.05in}
\begin{tabular}{|c|c|c|c|}
\hline
Dataset 
  & Support & Refute & NEI \\  \hline
  T-FEVER & 11,799 & 9,292 & 3,975  \\ \hline
  T-FEVEROUS & 33,357 & 28,959 & 1,266  \\ \hline
  T-QuanTemp & 1,261 & 3,470 & 1,349  \\ \hline
 \end{tabular} 
\end{center}
\vspace*{-0.15in}
\caption{Dataset characteristics. }
\vspace*{-0.15in}
\label{table:statistics2}
 \end{table}

\smallskip
\noindent\textbf{Implementation Details.}
We implement the \texttt{ChronoFact} framework using Hugging Face Transformers Library with PyTorch. The Event Encoder use  flan-T5 base \cite{flant5} with a hidden size  of 768. In the Multi-level Attention Encoder, the token-level, event-level, and time-level representations  pass through a linear layer of dimension 768 to calculate  attention scores. The hidden size of the fully connected layers is set to 192. The Chronological Order Classifier uses two  layers of Bi-LSTM, each with a hidden size of 768, and  the fully connected layers have a hidden size of  192, matching those in the  Claim Classifier.

We train the model using Adafactor with a batch size of 8 and a learning rate of 5e-5 for 5 epochs on each dataset. 
ChronoFact is trained to predict REF and SUP labels on ChronoClaims, as the dataset does not have NEI  labels.  
For T-FEVER and T-FEVEROUS, the model is trained to predict SUP, REF, and NEI.
We report the macro F1 score of the best performing model on the test set.

\begin{table*}[t!]
\begin{center}
\scriptsize
\begin{tabular}{|c||c||ccc||cc||}
\hline
\multirow{2}{*}{\textbf{ Method} } & \multirow{2}{*}{\textbf{ Overall } } & \multicolumn{3}{c||}{\textbf{Number of Events}} & \multicolumn{2}{c||}{\textbf{Event Types}} \\ \cline{3-7}
    & & {3 events} & {4 events} & {5 events}  & Overlapping & Recurring  \\ \hline
KGAT & 50.56$\pm$3.52 & 52.61$\pm$3.20 & 49.33$\pm$4.06 & 47.17$\pm$5.57 & 50.14$\pm$3.88 & 47.43$\pm$1.68 \\ \hline
CGAT & 56.69$\pm$0.36 & 58.11$\pm$1.20 & 55.98$\pm$0.57 & 54.75$\pm$0.58 & 56.56$\pm$0.22 & 56.95$\pm$0.97 \\ \hline
ITR & 58.34$\pm$6.44 & 63.86$\pm$3.49 & 61.52$\pm$0.27 & 51.15$\pm$2.41 & 57.87$\pm$5.25 & 55.50$\pm$6.31 \\ \hline
FAITH & 60.84$\pm$0.11 & 61.64$\pm$0.89 & 61.48$\pm$0.30 & 59.03$\pm$1.31 & 61.03$\pm$0.21 & 52.42$\pm$0.16  \\ \hline
GPT4o & 61.93$\pm$0.30 & 70.06$\pm$0.33 & 57.65$\pm$0.54 & 55.51$\pm$0.48 & 61.86$\pm$0.22 & 55.08$\pm$0.18 \\ \hline
TACV & 63.42$\pm$0.60 & 67.45$\pm$0.87 & 63.02$\pm$0.78 & 61.81$\pm$0.56 & 62.37$\pm$0.69 & 65.84$\pm$5.04 \\ \hline
\texttt{ChronoFact} & \textbf{85.49$\pm$0.53} & \textbf{86.33$\pm$0.46} & \textbf{85.55$\pm$0.31} & \textbf{84.86$\pm$0.43} & \textbf{84.86$\pm$0.76} & \textbf{80.89$\pm$0.55} \\ \hline
\end{tabular} 
\vspace*{-0.1in}
\caption{Comparison of macro F1 on ChronoClaims.}
\label{tab:comparative-chrono-macro}
\end{center}
\vspace*{-0.2in}
\end{table*}

\begin{table*}[t!]
\begin{center}
\scriptsize
\begin{tabular}{|c||c||ccc||cc||}
\hline
\multirow{2}{*}{\textbf{ Method} } & \multirow{2}{*}{\textbf{ Overall } } & \multicolumn{3}{c||}{\textbf{Number of Events}} & \multicolumn{2}{c||}{\textbf{Event Types}} \\ \cline{3-7}
    & & {3 events} & {4 events} & {5 events}  & Overlapping & Recurring  \\ \hline
KGAT & 53.12$\pm$1.54 & 54.81$\pm$2.21 & 52.63$\pm$1.41 & 51.47$\pm$1.02 & 52.50$\pm$0.10 & 51.27$\pm$0.98  \\ \hline
CGAT & 56.77$\pm$0.41 & 58.22$\pm$1.26 & 56.19$\pm$0.58 & 55.56$\pm$0.14 & 56.71$\pm$0.35 & 55.11$\pm$0.74  \\ \hline
ITR & 62.15$\pm$0.46 & 68.15$\pm$3.02 & 61.86$\pm$0.96 & 54.38$\pm$2.70 & 61.72$\pm$1.46 & 59.35 $\pm$2.66 \\ \hline
FAITH & 61.08$\pm$0.60 & 61.95$\pm$0.81 & 61.71$\pm$0.28 &  59.19$\pm$1.25 & 61.21$\pm$0.20 & 52.47$\pm$0.17  \\ \hline
GPT-4o in 0-shot & 62.00$\pm$0.30 & 70.20$\pm$0.32 & 57.89$\pm$0.51 & 56.25$\pm$0.46 & 61.91$\pm$0.21 & 55.51$\pm$0.18  \\ \hline
TACV & 65.22$\pm$0.71 & 67.79$\pm$0.91 & 64.23$\pm$0.73 & 63.01$\pm$0.63 & 63.70$\pm$0.75 & 61.09$\pm$0.80 \\ \hline
\texttt{ChronoFact} & \textbf{85.45$\pm$0.74} & \textbf{86.00$\pm$0.80} & \textbf{85.40$\pm$0.61} & \textbf{84.77$\pm$0.86} & \textbf{83.86$\pm$0.79} & \textbf{83.08$\pm$0.57} \\ \hline
\end{tabular} 
\vspace*{-0.1in}
\caption{Comparison of micro F1 on ChronoClaims.}
\label{tab:comparative-chrono}
\end{center}
\vspace*{-0.2in}
\end{table*}

\begin{table*}[t!]
\begin{center}
\scriptsize
\begin{tabular}{|c||c|cc||c|ccc||c||}
\hline
\multirow{3}{*}{\textbf{ Method} } & \multicolumn{3}{c||}{\textbf{T-FEVER}} & \multicolumn{4}{c||}{\textbf{T-FEVEROUS}} & \textbf{T-QuanTemp} \\ \cline{2-9}
    & \multirow{2}{*}{\textbf{ Overall } } & \multicolumn{2}{|c||}{\textbf{Number of Events}} &  \multirow{2}{*}{\textbf{ Overall } } & \multicolumn{3}{|c||}{\textbf{Number of Events}} & \multirow{2}{*}{\textbf{ Overall } } \\ \cline{3-4} \cline{6-8}
    & & 1 event & 2 events & & 1 event & 2 event & 3 event & \\ \hline
KGAT & 40.66$\pm$1.04 & 40.82$\pm$1.21 & 37.85$\pm$2.40 & 17.66$\pm$1.21 & 21.95$\pm$3.03 & 17.85$\pm$1.39 & 16.40$\pm$0.82 & 39.82$\pm$0.46 \\ \hline
CGAT & 42.31$\pm$2.47 & 42.47$\pm$2.61 & 39.20$\pm$0.85 & 19.62$\pm$1.92 & 23.78$\pm$1.66 & 19.30$\pm$1.85 & 18.54$\pm$1.95 & 42.01$\pm$0.25 \\ \hline
ITR & 45.21$\pm$3.86 & 45.62$\pm$3.89 & 36.77$\pm$2.36 & 27.62$\pm$2.99 & 30.19$\pm$4.50 & 27.64$\pm$3.39 & 26.99$\pm$2.55 & 44.90$\pm$1.27 \\ \hline
FAITH & 49.03$\pm$0.79 & 49.42$\pm$0.57 & 42.65$\pm$4.06 & 38.63$\pm$0.02 & 41.89$\pm$0.15 & 40.17$\pm$0.27 & 37.75$\pm$0.15 & 49.65$\pm$0.24 \\ \hline
GPT4o & 53.77$\pm$0.23 & 53.54$\pm$0.30 & 57.19$\pm$1.29 & 43.54$\pm$0.19 & 44.28$\pm$0.48 & 44.11$\pm$0.22 & 43.00$\pm$0.27 & 53.12$\pm$0.21 \\ \hline
TACV & 49.86$\pm$0.40 & 50.25$\pm$0.41 & 43.64$\pm$0.01 & 39.84$\pm$0.60 & 43.72$\pm$0.84 & 39.78$\pm$0.93 & 37.72$\pm$0.82 & 49.69$\pm$0.47 \\ \hline
\texttt{ChronoFact} & \textbf{56.29$\pm$1.50} & \textbf{56.14$\pm$1.41} & \textbf{57.34$\pm$2.39} & \textbf{47.78$\pm$0.98} & \textbf{48.27$\pm$1.62} & \textbf{48.23$\pm$2.41} & \textbf{47.88$\pm$2.21} & \textbf{65.67$\pm$0.20} \\ \hline
\end{tabular} 
\vspace*{-0.1in}
\caption{Comparison of macro F1 on T-FEVER, T-FEVEROUS, and T-QuanTemp.} 
\label{tab:comparative-tfever-tfeverous-macro}
\end{center}
\vspace*{-0.15in}
\end{table*}

\begin{table*}[t!]
\begin{center}
\scriptsize
\begin{tabular}{|c||c|cc||c|ccc||c||}
\hline
\multirow{3}{*}{\textbf{ Method} } & \multicolumn{3}{c||}{\textbf{T-FEVER}} & \multicolumn{4}{c||}{\textbf{T-FEVEROUS}} & \textbf{T-QuanTemp} \\ \cline{2-9}
    & \multirow{2}{*}{\textbf{ Overall } } & \multicolumn{2}{|c||}{\textbf{Number of Events}} &  \multirow{2}{*}{\textbf{ Overall } } & \multicolumn{3}{|c||}{\textbf{Number of Events}} & \multirow{2}{*}{\textbf{ Overall } } \\ \cline{3-4} \cline{6-8}
    & & 1 event & 2 events & & 1 event & 2 event & 3 event & \\ \hline
KGAT & 43.30$\pm$0.99 & 43.49$\pm$1.05 & 40.80$\pm$1.84 & 16.79$\pm$0.99 & 19.20$\pm$2.00 & 16.40$\pm$0.78 & 14.62$\pm$1.37 & 57.49$\pm$0.38 \\ \hline
CGAT & 44.72$\pm$1.59 & 44.87$\pm$1.68 & 43.15$\pm$0.37 & 19.13$\pm$2.29 & 21.71$\pm$2.27 & 18.39$\pm$2.32 & 16.76$\pm$2.64 & 58.35$\pm$0.33 \\ \hline
ITR & 45.85$\pm$2.14 & 46.24$\pm$1.99 & 37.61$\pm$1.44 & 33.11$\pm$1.25 & 34.23$\pm$1.93 & 33.51$\pm$1.54 & 31.48$\pm$1.55 & 58.20$\pm$1.47 \\ \hline
FAITH & 51.75$\pm$0.57 & 52.07$\pm$0.40 & 46.57$\pm$3.16 & 51.51$\pm$0.06  & 52.65$\pm$0.21 & 51.28$\pm$0.51 & 49.98$\pm$0.13  & 59.11$\pm$0.21 \\ \hline
GPT-4o in 0-shot & 55.28$\pm$0.26 & 55.21$\pm$0.35 & 57.69$\pm$1.11 & 53.01$\pm$0.09 & 52.66$\pm$0.19 & 53.81$\pm$0.34 & 52.94$\pm$0.26 & 62.70$\pm$0.12  \\ \hline
TACV & 52.53$\pm$0.64 & 52.77$\pm$0.66 & 48.72$\pm$0.80 & 53.77$\pm$0.21 & 54.92$\pm$0.53 & 54.86$\pm$0.35 & 52.45$\pm$0.63 & 60.52$\pm$0.21 \\ \hline
\texttt{ChronoFact} & \textbf{61.58$\pm$0.49} & \textbf{61.75$\pm$0.55} & \textbf{60.04$\pm$0.74} & \textbf{58.37$\pm$0.32} & \textbf{59.16$\pm$0.62} & \textbf{58.70$\pm$0.31} & \textbf{58.23$\pm$0.23} & \textbf{70.45$\pm$0.33}  \\ \hline
\end{tabular} 
\vspace*{-0.1in}
\caption{Comparison of micro F1 on T-FEVER, T-FEVEROUS, and T-QuanTemp.} 
\label{tab:comparative-tfever-tfeverous}
\end{center}
\vspace*{-0.15in}
\end{table*}

\subsection{Sensitivity Experiments}

\begin{figure}[t!] 
  \vspace*{-0.1in}
  \includegraphics[height= 1.2in,width=.23\textwidth]{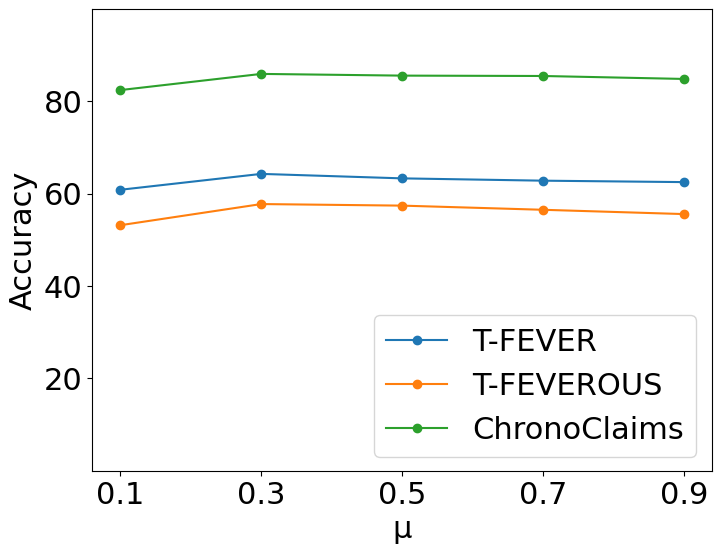}
  \hfill
  \includegraphics[height= 1.2in,width=.23\textwidth]{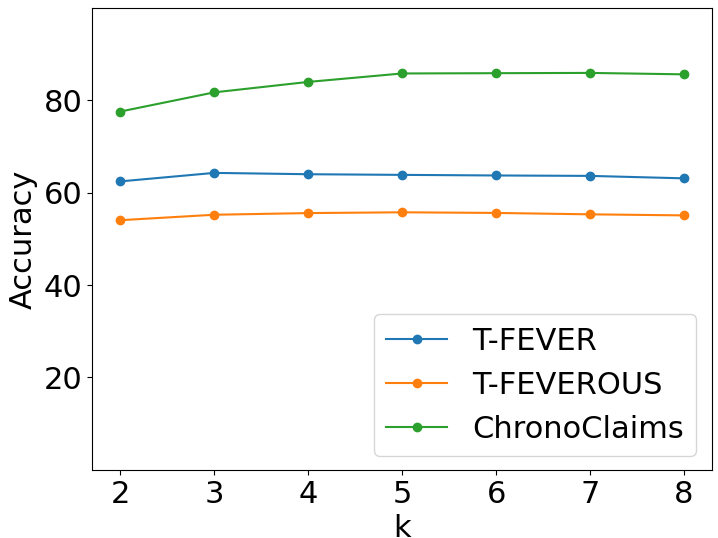}
    \vspace{-0.1in}
   \hspace*{0.5in} (a) Effect on $\mu$ \hspace*{0.9in} (b) Effect on $k$
\caption{Sensitivity Experiments on \texttt{ChronoFact}.}
\label{fig:scores-mu-variation}
\vspace*{-0.1in}
\end{figure}

We first conduct sensitivity experiments to obtain the optimal value of the parameter $\mu$ in  Eqn~\ref{eq:loss_final}. \autoref{fig:scores-mu-variation}(a) shows the label accuracy  as we vary the 
value of $\mu$ from 0.1 to 0.9. 
The performance improves when $\mu$ increases from 0.1 to 0.3, indicating that the model benefits from incorporating $\mathcal{L}_{soft}$. The best  performance is achieved when  $\mu = 0.3$ across all datasets and we use this value for the rest of the experiments.

We also vary the number of top-$k$ evidence events in the classifier module to examine its effect on \texttt{ChronoFact}'s performance. \autoref{fig:scores-mu-variation}(b) shows that the  optimal performance for T-FEVER, T-FEVEROUS, and \texttt{ChronoClaims} was achieved when  $k$ is 3, 5, and 7 respectively, and we use these values in our experiments.
Note that ChronoClaims generally achieves higher performance because it has only two class labels (SUPPORTS, REFUTES), whereas T-FEVER and T-FEVEROUS include an additional label ('NOT ENOUGH INFO').

\subsection{Comparative Experiments}
We compare \texttt{ChronoFact} with the following state-of-the-art evidence-based fact verification baselines:
    
    $\bullet$ KGAT \cite{KGAT}. This  uses a transformer to obtain claim-sentence representations and a graph attention network  to aggregate the evidence for claim verification.

    $\bullet$ CGAT \cite{cgat}. This method incorporates  external knowledge from ConceptNet to enrich the contextual representations of claim and evidence sentences. It then employs graph attention networks to propagate the information among the evidence sentences to verify the claim veracity.

  $\bullet$ ITR  \cite{implicit-temporal-reasoning}. This also employs a transformer to obtain the claim and evidence representations which 
  are augmented with the publication dates of the claim and evidence for temporal reasoning.

  $\bullet$ FAITH \cite{faith}. We adapt this temporal QA model for temporal claim verification by using GPT to generate relevant temporal questions for FAITH, then prompting GPT to verify the claim based on FAITH's generated answers.

  $\bullet$ GPT-4o in a zero-shot setting. Given the claim and retrieved evidence using TACV retrieval method, we prompt GPT-4o to predict the claim's label.

  $\bullet$ TACV \cite{tacv}. This is an end-to-end solution for temporal claim verification that considers the temporal information in claims to obtain relevant evidence sentences and uses LLM for temporal reasoning.

\autoref{tab:comparative-chrono-macro} shows the macro F1 score of the various methods on the ChronoClaims dataset, with our  proposed \texttt{ChronoFact} achieving the highest score and significantly outperforming  the baseline models. 
Further analysis in terms of the number of events shows that \texttt{ChronoFact} is better in managing complex claims with multiple events. This is  evident in the smaller decline in the performance as the number of events increases, compared to other baselines.
We also analyze \texttt{ChronoFact}'s ability to handle complex temporal event types where the timelines of different events may overlap or the events may recur at multiple time points.
 Once again, there is a significant gap in the macro F1 score compared to other methods, demonstrating \texttt{ChronoFact}'s superior performance in managing such complexities.

\autoref{tab:comparative-tfever-tfeverous-macro} shows the  performance on T-FEVER and T-FEVEROUS datasets, where 
 \texttt{ChronoFact} is the best performer on both datasets. 
 Similar trend is observed when comparing the performance in terms of the number of events.
\autoref{tab:comparative-tfever-tfeverous-macro} also shows \texttt{ChronoFact}'s robustness on the real-world T-QuanTemp, despite noisy date information.

\begin{table*}[t!]
\begin{center}
\scriptsize
\begin{tabular}{|c|c|c|c|c|}
\hline
\multirow{2}{*}{\textbf{Variants}}  & \multicolumn{4}{c|}{\textbf{Datasets}}\\ \cline{2-5} 
& ChronoClaims & T-FEVER  & {T-FEVEROUS} & {T-QuanTemp}   \\ \hline
 w/o multi-level attention encoder   &  84.65$\pm$0.30 & 53.60$\pm$0.91 & 42.01$\pm$0.55 & 63.53$\pm$1.51 \\ \hline
 w/o claim event classifier & 83.72$\pm$0.57 & 54.55$\pm$1.10 & 41.24$\pm$2.28 & 63.95$\pm$0.98 \\ \hline
w/o chronological order classifier & 76.88$\pm$2.14 & 55.31$\pm$0.32 & 40.75$\pm$1.14 & 64.35$\pm$0.43 \\ \hline
\texttt{ChronoFact} &  \textbf{85.49$\pm$0.53} &  \textbf{56.29$\pm$1.50} &  \textbf{47.78$\pm$0.98} & \textbf{65.67$\pm$0.20} \\ \hline
\end{tabular}\\
\vspace*{-0.1in}
\caption{Macro F1 score of ablation studies.}

\label{table:ablation-studies-macro}
\end{center}
\vspace*{-0.1in}
\end{table*}

\begin{table*}[t!]
\begin{center}
\scriptsize
\begin{tabular}{|c|c|c|c|c|}
\hline
\multirow{2}{*}{\textbf{Variants}}  & \multicolumn{4}{c|}{\textbf{Datasets}}\\ \cline{2-5} 
& ChronoClaims & T-FEVER  & {T-FEVEROUS} & {T-QuanTemp}   \\ \hline
 w/o multi-level attention encoder   &  84.87$\pm$0.32 & 60.64$\pm$0.08 & 55.69$\pm$1.03 & 69.52$\pm$1.23 \\ \hline
 w/o claim event classifier & 83.83$\pm$0.50 & 60.78$\pm$0.59 & 55.25$\pm$0.77 & 70.12$\pm$0.27 \\ \hline
w/o chronological order classifier & 79.70$\pm$1.54 & 61.18$\pm$0.39 & 53.80$\pm$1.00 & 70.30$\pm$0.40 \\ \hline
\texttt{ChronoFact} &  \textbf{85.45$\pm$0.74} &  \textbf{61.58$\pm$0.49} &  \textbf{58.37$\pm$0.32} & \textbf{70.45$\pm$0.33} \\ \hline
\end{tabular}\\
\vspace*{-0.1in}
\caption{Micro F1 score of ablation studies.}

\label{table:ablation-studies-micro}
\end{center}
\vspace*{-0.1in}
\end{table*}

\smallskip
\noindent\textbf{Error Analysis.}
We conduct an error analysis on 50 randomly incorrectly predicted claims on ChronoClaim dataset.
We find that 82\% of these errors are due to claims containing implicit temporal information.
For example, the claim "Greg Clark first held the position of Minister of State for Decentralisation, thereafter became the Secretary of State for Business and Trade, and later served as the Secretary of State for Levelling Up, Housing and Communities." lacks explicit dates or clear temporal markers.
This ambiguity poses a challenge for the model to accurately encode and reason about the sequence of events. 
The remaining 18\% of the errors comes from claims with events that have multiple dates and varying levels of granularity, such as "Nicola Fratoianni joined the Movement for the Left from 2009 until October 22, 2010". 
The complexity of this claims hinders the model's ability to  encode the temporal information.

\subsection{Ablation Studies}

We examine the effectiveness of the key modules in
\texttt{ChronoFact} by
implementing the following  variants:

$\bullet$ \texttt{ChronoFact} without multi-level attention encoder module. In this variant,  we set the final attention scores between each claim event $c_i$ and evidence event $e_j$ to 1.

$\bullet$ \texttt{ChronoFact} without claim event classifier. Here, the input to the claim classifier is  the concatenation of $seqC$, $seqE$, and the  distribution of chronological order  $z^o$.

$\bullet$ \texttt{ChronoFact} without chronological order classifier. This variant
does not assess the consistency of the 
 chronological order of the claim events with that of the evidence events. Therefore, the input to the Claim Classifier is   the concatenation of $seqC$, $seqE$, and the probability distribution of the claim events $z^{c_1} \cdots z^{c_n}$.

\autoref{table:ablation-studies-macro} shows that 
the largest performance drop occurs when 
 when we exclude the chronological order classifier, emphasizing that predicting chronological order enhances the model prediction. This effect is particularly evident in the ChronoClaims dataset, which is specifically designed to test the model's ability to reason over the chronological order of events.
Its effect is less noticeable on the T-FEVER dataset as it mainly consists of single-event claims.
The next largest drop in macro F1 score is when we exclude the claim event classifier, highlighting the importance of predicting individual claim events for the overall claim prediction.
 
Excluding the multi-level attention encoder leads to a decrease in  macro F1 score across all datasets, indicating the role of considering the relevance of evidence events in claim verification. 
We conduct a manual analysis of the attention scores associated with the relevant evidence events  for 50 claim events.
For each claim event, we examine whether the evidence event with the highest event-level and token-level attention scores is semantically related, and whether the evidence event with the highest time-level attention score is temporally related. 
 Our findings indicate that, in every case, the evidence event with the highest score is indeed semantically or temporally relevant to the corresponding claim event.

\section{Case Studies}

Finally, we present case studies to illustrate the importance of considering the chronological order in verifying temporal claims.
\autoref{table:case-studies-feverous} shows a  T-FEVEROUS claim with three events: "Yossi Yona studying for a PhD", "Yossi Yona become a professor at Ben-Gurion University", and "Yossi Yona join the left camp of Israel party".
The temporal context provided by the word "before" is lost between the events "Yossi Yona became a Professor" and "Yossi Yona joined the Left Camp of Israel Party", resulting in each individual claim event being supported by the retrieved evidence.
However, the chronological order of the claim events is incorrect because evidence shows that "Yossi Yona joined the left camp of Israel party" \textit{before} "became a professor".
\texttt{ChronoFact} addresses this issue by  inferring the missing temporal relationships
from the timeline of events, and
correctly predicts the label as REF.
In contrast, TACV evaluates  the claim events independently and  mistakenly predicts the claim as SUP.

\begin{table*}[t!]
\vspace*{-0.15in}
\begin{center}
\small
\begin{tabular}{| p{0.1\textwidth} | p{0.65\textwidth}|p{0.055\textwidth}|p{0.05\textwidth}|p{0.05\textwidth}|}
\hline
\multicolumn{5}{|p{0.97\textwidth}|}{\textbf{Claim:} Yossi Yona began studying for a PhD and went on to become a Professor of philosophy of education at Ben - Gurion University before he joined the Left Camp of Israel party . \hfill Ground Truth Label: REF} \\ \hline

 \multirow{3}{*}{Claim Events} 
& {c1: Yossi Yona began studying for a PhD } &  {Claim } & {Chrono.}&{Final} \\
& {c2: Yossi Yona  become a Professor of philosophy of education at Ben - Gurion University} & Event  & Order & Claim  \\
& {c3: Yossi Yona joined the Left Camp of Israel party }  &Label &  & \\ \hline

{TACV} & \textbf{Evidence Events:} & & &\\
 & {\textbullet{ After graduating in 1979, he began studying for a PhD, graduating from the University of Pennsylvania in Philadelphia}}  & { SUP c1} & N.A. & SUP \\ 
 & {\textbullet~{Yossi Yona became a Professor of philosophy of education at Ben-Gurion University} \newline \textbullet~{He subsequently joined the Education Department at Ben-Gurion University of the Negev.}}   & SUP c2 &  &  \\ 
 & {\textbullet~{Whilst at university he joined the Left Camp of Israel party.}}  & SUP c3&  &   \\ \hline 

{\texttt{ChronoFact}} & \textbf{Evidence Events in Chronological Order:}& & & \\
 & {1. Whilst at university, Yossi Yona joined the Left Camp of Israel party} & SUP c3& REF & REF \\
 & {2. After graduating in 1979, Yossi Yona began studying for a PhD, graduating from the University of Pennsylvania in Philadelphia} \newline 3. {He subsequently joined the Education Department at Ben-Gurion University of the Negev.}   & SUP c1&  &  \\
  & 4. {Yossi Yona became a Professor of philosophy of education at Ben-Gurion University}  & SUP c2 &  &  \\ \hline

\end{tabular}
\vspace*{-0.1in}
\caption{Sample T-FEVEROUS claim where chronological order of claim events is inconsistent with that of evidence events. }
\label{table:case-studies-feverous}
\end{center}
\vspace*{-0.1in }
\end{table*}

\begin{table*}[t!]
\begin{center}
\small
\begin{tabular}{| p{0.11\textwidth} | p{0.6\textwidth}|p{0.055\textwidth}|p{0.055\textwidth}|p{0.055\textwidth}|}
\hline

\multicolumn{5}{|p{0.97\textwidth}|}{\textbf{Claim:} Davie Dodds was a member of Dundee United F.C. before joining Arbroath F.C., thereafter becoming part of the Scotland national football team, then moving to Neuchâtel Xamax, and finally playing for Aberdeen F.C. \hfill Ground Truth Label: SUP} \\ \hline
 \multirow{3}{*}{Claim Events} 
& {c1: Davie Dodds was a member of Dundee United F.C. } &  {Claim } & {Chrono.}&{Final} \\
& c2: Davie Dodds joined Arbroath F.C. after joining Dundee United F.C. & Event  & Order & Claim  \\
& {c3: Davie Dodds became part of the Scotland national football team }  & Label &  &\\
& {c4: Davie Dodds moved to Neuchâtel Xamax} & & & \\
& {c5: Davie Dodds played for Aberdeen F.C.} & & & \\ \hline

{TACV} & \textbf{Evidence Events:} & & &\\
 & \textbullet~{Davie Dodds is a member of the Dundee United F.C. from 1975 until 1986}
 & SUP c1 & N.A. & REF \\
&  \textbullet~{Davie Dodds is a member of the Arbroath F.C. from 1977 until 1978}  & {REF c2} & &  \\ 
 &\textbullet~{Davie Dodds is a member of the Scotland national football from 1983 until 1983.}  & SUP c3 &  &  \\ 
 & \textbullet~{Davie Dodds is a member of the Neuchâtel Xamax from 1986 until 1986}  & SUP c4&  &   \\ 
 & \textbullet~{Davie Dodds is a member of the Aberdeen F.C. from 1986 until 1989} & SUP c5& & \\
 \hline 

{\texttt{ChronoFact}} & \textbf{Evidence Events in Chronological Order:}& & & \\
 & {1. Davie Dodds is a member of the Dundee United F.C. from 1975 until 1986} & SUP c1& SUP & SUP\\
 & {2. Davie Dodds is a member of the Arbroath F.C. from 1977 until 1978}& SUP c2&  &\\
& 3. {Davie Dodds is a member of the Scotland national football from 1983 until 1983}   & SUP c3&  &  \\
  & 4. {Davie Dodds is a member of the Neuchâtel Xamax from 1986 until 1986}  & SUP c4 &  &  \\ 
   & 5. {Davie Dodds is a member of the Aberdeen F.C.  from 1986 until 1989}  & SUP c5 &  & \\
  
  \hline

\end{tabular}
\vspace*{-0.1in}
\caption{Sample claim in ChronoClaims dataset involving overlapping events. }
\label{table:case-studies-overlap}
\end{center}
\vspace*{-0.2in}
\end{table*}

\begin{table*}[t!]
\begin{center}
\small
\begin{tabular}{| p{0.1\textwidth} | p{0.65\textwidth}|p{0.055\textwidth}|p{0.05\textwidth}|p{0.04\textwidth}|}
\hline
\multicolumn{5}{|p{0.97\textwidth}|}{\textbf{Claim:} Georgi Andonov was a member of the Botev Plovdiv from 2002 to 2006, joined the Bulgaria national under-21 team from 2003 to 2005, returned to Botev Plovdiv in 2009, played for PSFC Chernomorets Burgas from 2010 to 2012, and then became a member of PFC Beroe Stara Zagora starting in 2015. \hfill Ground Truth Label: SUP} \\ \hline
 \multirow{3}{*}{Claim Events} 
& {c1: Georgi Andonov was a member of the Botev Plovdiv from 2002 to 2006 } &  {Claim } & {Chrono.}&{Final} \\
& c2: Georgi Andonov joined the Bulgaria national under-21 team from 2003 to 2005. & Event  & Order & Claim  \\
& {c3: Georgi Andonov returned to Botev Plovdiv in 2009 }  & Label &  &\\
& {c4: Georgi Andonov played for PSFC Chernomorets Burgas from 2010 to 2012} & & & \\
& {c5: Georgi Andonov became a member of PFC Beroe Stara Zagora starting in 2015} & & & \\ \hline

{TACV} & \textbf{Evidence Events:} & & &\\
& \textbullet { Georgi Andonov is a member of the Botev Plovdiv from 2002 until 2006} 
 &  SUP c1 & N.A. & REF \\
 &  & REF c3 && \\
 & \textbullet { Georgi Andonov is a member of the Bulgaria national under-21 team from 2003 until 2005 }  & SUP c2 &  &  \\ 
 & \textbullet { Georgi Andonov is a member of the PSFC Chernomorets Burgas from 2010 until 2012 }  & SUP c4&  &   \\ 
 & \textbullet { Georgi Andonov is a member of the PFC Beroe Stara Zagora from 2015} & SUP c5& & \\
 &\textbullet { Georgi Andonov is member of Botev Plovdiv from 2009 until 2009 }  &&&\\
 \hline 

{\texttt{ChronoFact}} & \textbf{Evidence Events in Chronological Order:}& & & \\
 &  {1. Georgi Andonov is a member of the Botev Plovdiv from 2002 until 2006} & SUP c1& SUP & SUP\\
 & {2. Georgi Andonov is a member of the Bulgaria national under-21  team from 2003 until 2005}& SUP c2&  &\\
& 3. {Georgi Andonov is a member of the Botev Plovdiv from 2009 until 2009}   & SUP c3&  &  \\
  & 4. {Georgi Andonov is a member of the PSFC Chernomorets Burgas from 2010 until 2012}  & SUP c4 &  &  \\ 
   & 5. {Georgi Andonov is a member of the PFC Beroe Stara Zagora from 2015}  & SUP c5 &  & \\ \hline
\end{tabular}
\vspace*{-0.1in}
\caption{Sample claim in ChronoClaims dataset involving recurring event. }
\label{table:case-studies-recurring}
\end{center}
\vspace*{-0.1in}
\end{table*}

\autoref{table:case-studies-overlap} shows a claim from ChronoClaim dataset. The claim has five events, where the first event "Davie was a member of Dundee United F.C"  overlaps with the second event "Davie joined Arbroath F.C.". 
TACV  does not accommodate the possibility of overlapping events and assumes  that "join" implies a full transfer or departure, and conclude that the evidence refutes the claim event c2.
 In contrast, \texttt{ChronoFact} can handle overlapping events and correctly conclude that c2 is supported by the evidence.

\autoref{table:case-studies-recurring} shows another claim which involves a recurring event where Georgi Andonov was a member of Botev Polediv from 2002 to 2006, left, and rejoined in 2009.
TACV fails to ecognize this recurrence, mistakenly using evidence of Andonov's initial membership period (2002-2006) to refute the claim of his 2009 return.
In contrast, \texttt{ChronoFact} organizes the evidence chronologically, recognizing Georgi Andonov’s return to Botev Plovdiv in 2009 after a hiatus, 
yielding a correct prediction.

\section{Conclusion}

We have introduced  a framework for temporal claim verification that incorporates temporal reasoning based on the chronological order of events.  By utilizing a multi-level attention encoder, \texttt{ChronoFact} effectively captures the relevance of  evidence events to  claim events. This enables \texttt{ChronoFact} to accurately predict to claim event labels and verify the chronological order consistency between claim and evidence events before determining the overall claim label.
We have curated a new benchmark dataset  that involves  complex claims with multiple events that may overlap or recur. Extensive experiments  on multiple datasets have shown that \texttt{ChronoFact} outperforms state-of-the-art models.

\appendix

\section*{Acknowledgments}
This work is supported by the Ministry of Education, Singapore, under its MOE AcRF TIER 3 Grant (MOE-MOET32022-0001).

\bibliographystyle{named}
\bibliography{ijcai25}

\end{document}